\documentclass[letterpaper, 10 pt, conference]{ieeeconf}  

\IEEEoverridecommandlockouts    

\let\labelindent\relax
\usepackage{enumitem}
\usepackage{booktabs}
\usepackage{svg} 
\usepackage{amsmath} 
\usepackage{graphicx}
\usepackage{subfigure}
\usepackage{parskip}
\usepackage{algorithm}
\usepackage[noend]{algpseudocode}
\usepackage{stfloats}
\usepackage{caption}
\usepackage{setspace}
\usepackage{cite}
\usepackage[bookmarks=true, hidelinks]{hyperref}
\usepackage{lipsum}
\usepackage{multirow}
\usepackage{booktabs}  
\usepackage{threeparttable}  
\usepackage{bbding}
\usepackage{arydshln}
\usepackage{colortbl}
\usepackage{graphics} 
\usepackage{epsfig} 
\usepackage{times} 
\usepackage{amssymb}

\graphicspath{ {./images/} }

\hypersetup{hidelinks,
	        colorlinks=true,
	        allcolors=blue,
	        pdfstartview=Fit,
	        breaklinks=true}

\title{\LARGE \bf
    RMBench: Benchmarking Deep Reinforcement Learning for Robotic Manipulator Control
}

\author{Yanfei Xiang$^{1}$, Xin Wang$^{2,\S}$, Shu Hu$^{3}$, Bin Zhu$^{4}$, Xiaomeng Huang$^{1}$, Xi Wu$^{5,\S}$, Siwei Lyu$^{2}$
\thanks{$^1$ Department of Earth System Science, Ministry of Education Key Laboratory for Earth System Modeling, Institute for Global Change Studies, Tsinghua University, Beijing 100084, China}%
\thanks{$^2$ University at Buffalo, State University of New York, USA}%
\thanks{$^3$ Carnegie Mellon University, USA}%
\thanks{$^4$ Microsoft Research Asia, China}%
\thanks{$^5$ Chengdu University of Information Technology, China}
\thanks{$^\S$ Corresponding authors: {xwang264}@buffalo.edu, xi.wu@cuit.edu.cn}%
}

\begin{document}

\captionsetup{font={small, stretch=1}}

\thispagestyle{empty}
\pagestyle{empty}

\maketitle

\begin{abstract}
Reinforcement learning is applied to solve actual complex tasks from high-dimensional, sensory inputs. The last decade has developed a long list of reinforcement learning algorithms. Recent progress benefits from deep learning for raw sensory signal representation. One question naturally arises: how well do they perform concerning different robotic manipulation tasks? Benchmarks use objective performance metrics to offer a scientific way to compare algorithms. In this paper, we present \textit{RMBench}, the first benchmark for robotic manipulations, which have high-dimensional continuous action and state spaces. We implement and evaluate reinforcement learning algorithms that directly use observed pixels as inputs. We report their average performance and learning curves to show their performance and stability of training. Our study concludes that none of the studied algorithms can handle all tasks well, soft Actor-Critic outperforms most algorithms in average reward and stability, and an algorithm combined with data augmentation may facilitate learning policies. Our code is publicly available at {\scriptsize\url{https://github.com/xiangyanfei212/RMBench-2022.git}}, including all benchmark tasks and studied algorithms. 


\end{abstract}

\section{Introduction}

Robotic manipulations refer to ways a robot (a robotic arm) interacts with objects around it, such as elevating an object above a threshold height, moving an end effector to a target location, and placing an object on top of another object. A robot must control its arms to accomplish tasks intelligently. Reinforcement Learning (RL) aims to learn a good strategy to maximize cumulative rewards for the agent by interacting with the environment. For robot manipulations, RL algorithms bring the hope of machines having human-like abilities by directly learning dexterous manipulations from observed raw pixels~\cite{c106}, including model-based approaches~\cite{c33, c34} and model-free approaches~\cite{c35, c36, c37, c38}.

Historically, model-based RL is applied to robotics first due to its high sample efficiency~\cite{c39}, allowing agents to complete tasks with a minimal number of iterations. Nevertheless, it can be difficult to learn precise models for complicated robotics controlling, leading to poor performance of model-based methods. Deep Deterministic Policy Gradient (DDPG)~\cite{c112} is developed to address problems with continuous state and action spaces, opening the door for other model-free RL algorithms to solve the robotic manipulation tasks with continuous state and action spaces. Since then, Twin Delayed Deep Deterministic Policy Gradient (TD3)~\cite{c114} and Soft Actor Critic (SAC)~\cite{c115} have been developed, leading to significant advancements in this field. Since robotic manipulation tasks have continuous states and actions, the sensory states (i.e. observations) usually carry meaningful physical attributes (position, velocity, force etc.) and vary over time. Impressive results have also been obtained using raw sensory inputs by combining with advances in deep learning~\cite{c100, c101, c102, c103}.

Advances in deep reinforcement learning have enabled autonomous agents to perform well on Atari games, often outperforming humans, using only raw pixels to make their decisions. However, most of these games take place in 2D environments that are fully observable to the agent. Because of the variations in shadow and light intensity, a robotic agent could not be able to fully comprehend the current state of the 3D environment from the 2D observed states. A critical insight in solving control tasks in real-world is learning better low-dimensional representations through autoencoders~\cite{c43} and data augmentations~\cite{c103, c122}. Therefore, it is necessary to find whether RL algorithms have human-like abilities by directly learning dexterous manipulation from observed raw pixels.

Benchmarks can provide a systematic evaluation of strengths and limitations of existing algorithms. The Arcade Learning Environment (ALE)~\cite{c104} provides a set of standard benchmarks for evaluating and comparing RL algorithms for tasks with high-dimensional state inputs and discrete actions. However, these benchmarks are unsuitable for comparing RL algorithms designed for tasks with continuous actions, such as robotic manipulations. Duan et al.~\cite{c105} propose a benchmark suite for continuous control tasks, including tasks with very high dimensional states and actions such as 3D humanoid locomotion. However, their benchmark is not designed for robotic manipulation tasks. It is well known that it is hard to develop continuous control algorithms that can observe high-dimensional images in RL, which are actively studied in robotic manipulations using reinforcement learning. It is crucial to develop a consistent and rigorous testbed to benchmark these algorithms to facilitate the study and improvement of robot manipulation solutions by providing a simulation framework and environment benchmark.

In this paper, we present a benchmark for RL learning-based robotic manipulations, called \textit{RMBench}, to evaluate the human-like abilities of RL algorithms by directly learning dexterous manipulations from observed raw pixels in the 3D environment. It consists of five types of representative robotic manipulation tasks, including lifting, placing, reaching, stacking, and reassembling. To evaluate RL algorithms with the benchmark, we implement the following state-of-the-art methods from the two categories (policy optimization and actor-critic): Vanilla Policy Gradient (VPG), Trust Region Policy Optimization (TRPO), Proximal Policy Optimization (PPO), Deep Deterministic Policy Gradient (DDPG), Delayed DDPG (TD3), and Soft Actor-Critic (SAC). Furthermore, we use DrQ-v2 involving autoencoders and data augmentation to improve environment learning. Considering vanilla policy gradient updates have no bias but high variance, we utilize an effective variance reduction scheme for policy gradients: General Advantage Estimation (GAE) in VPG, TRPO, and PPO. \textit{RMBench} uses observed raw pixels as input directly for each task to evaluate an algorithm's performance and stability in training policies. Based on \textit{RMBench} results, we conclude that none of the tested algorithms can handle all the tasks well. Overall, actor-critic methods outperform policy-optimization methods. VPG is the simplest algorithm but performs well in most tasks. Among the tested algorithms, SAC is effective for training deep neural network policies in terms of average reward and stability. Data augmentation techniques and autoencoder used in DrQ-v2 can help agents gain richer information about the environment, even when the learning is unstable.

Our main contributions can be summarized as follows:
\begin{itemize}[itemsep=0pt,topsep=0pt,parsep=0pt,leftmargin=*]
    \item We present the first benchmark for RL-based robotic manipulation tasks with high dimensional continuous states and actions. Five common types of environments are implemented, including lifting, placing, reaching, stacking, and reassembling.
    \item We provide a standardized set of RL-based benchmarking experiments as baselines for visual robotic manipulation tasks. Since that agent trained in a 3D simulation environment can be transferred to real-world tasks. We intend to facilitate the investigation and improvement of robot manipulation solutions.  
    \item Novel findings are presented based on the performance and learning stability. SAC is the most effective method. Although VPG is the simplest algorithm, it performs satisfactorily in most tasks. Data augmentation and autoencoder can help agents gain richer information about environments.
    
\end{itemize}

\section{Related Work}
 
Agents in many benchmarks, such as robosuite~\cite{c19}, ROBEL~\cite{c20}, and RLLib~\cite{c23}, receive low-dimensional physical states as input to the policy. Several RL competitions with low-dimensional actions have also been held~\cite{c24, c25}. In contrast, \textit{RMBench} contains tasks with high-dimensional continuous state and action spaces. The RL algorithms we implement and evaluate train policies using raw pixels as input directly.

Benchmarks for the Atari game with 2D state spaces have been proposed. For instance, ALE~\cite{c104} provides a benchmark for hundreds of Atari 2600 game environments. Liang et al.~\cite{c22} propose a benchmark to evaluate the importance of key representational biases encoded by DQN’s network and provides a generic representation for ALE, significantly reducing the burden of learning a representation for each game. Ankesh et al.~\cite{c21} systematically evaluate the importance of key representational biases encoded by DQN’s network by proposing simple linear representations. They provide a simple, reproducible benchmark, which is a generic representation for the ALE, significantly reducing the burden of learning a representation for each game. 
Advances in deep RL have enabled agents to perform well on Atari games. However, most of these games take place in 2D environments that are fully observable to the agent. A robotic agent cannot have full knowledge of the current state of the 3D environment. Therefore, building intelligent agents for real-world environments requires the capacity to capture an environment's latent characteristics. This paper concentrates on robotic manipulation tasks in 3D simulation environments.

Benchmarks on robotic manipulations can be based on real-world or simulated manipulations. Dasari et al.~\cite{c28} is a benchmark for the standard 2-finger gripper based on real-world manipulations. It includes four tasks: pouring, scooping, zipping, and insertion. They use the benchmark to compare five algorithms that can be divided into two categories: open-loop behavior cloning and a model-based offline RL algorithm (MOReL~\cite{c29}). The benchmark is insufficient for comparing current state-of-the-art RL algorithms. Simulation often provides a lower-cost alternative to sampling data from the real world. Aumjaud et al.~\cite{c30} propose a benchmark that trains policies in simulation before transferring them to an actual robotic manipulator. They apply the benchmark to compare the performance of various model-free RL algorithms by solving the reaching task with a robot manipulator. However, evaluation based on a single task type is insufficient to evaluate an algorithm's effectiveness thoroughly. In contrast, \textit{RMBench} implements five types tasks: lifting, placing, reaching, stacking, and reassembling.

\section{Our Robotic Manipulation Benchmark}

\begin{figure*}[h]
	\centering
	\setcounter {subfigure} {0} a){\includegraphics[width=105pt, height=90pt]{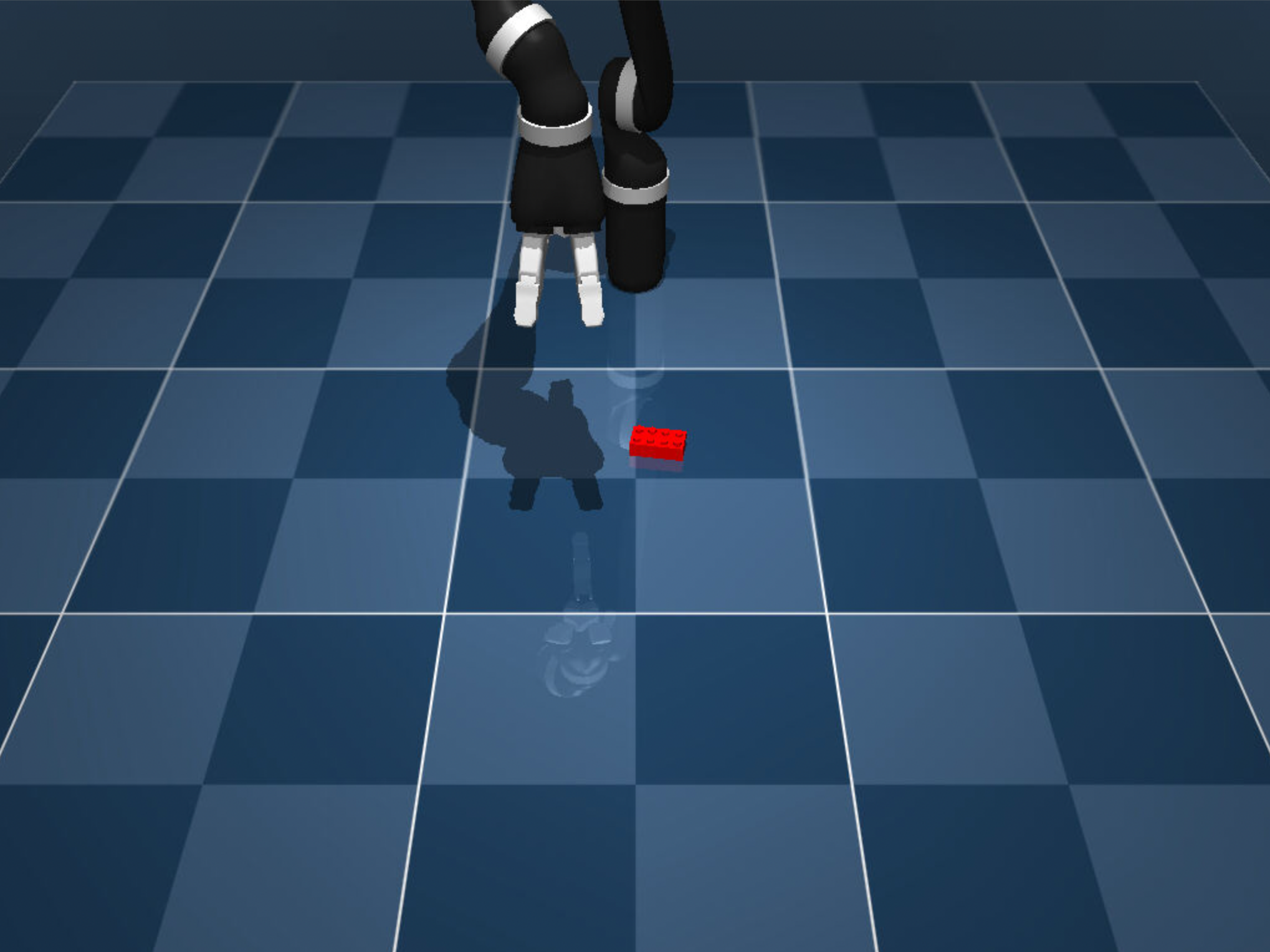}}
	\setcounter {subfigure} {0} b){\includegraphics[width=105pt, height=90pt]{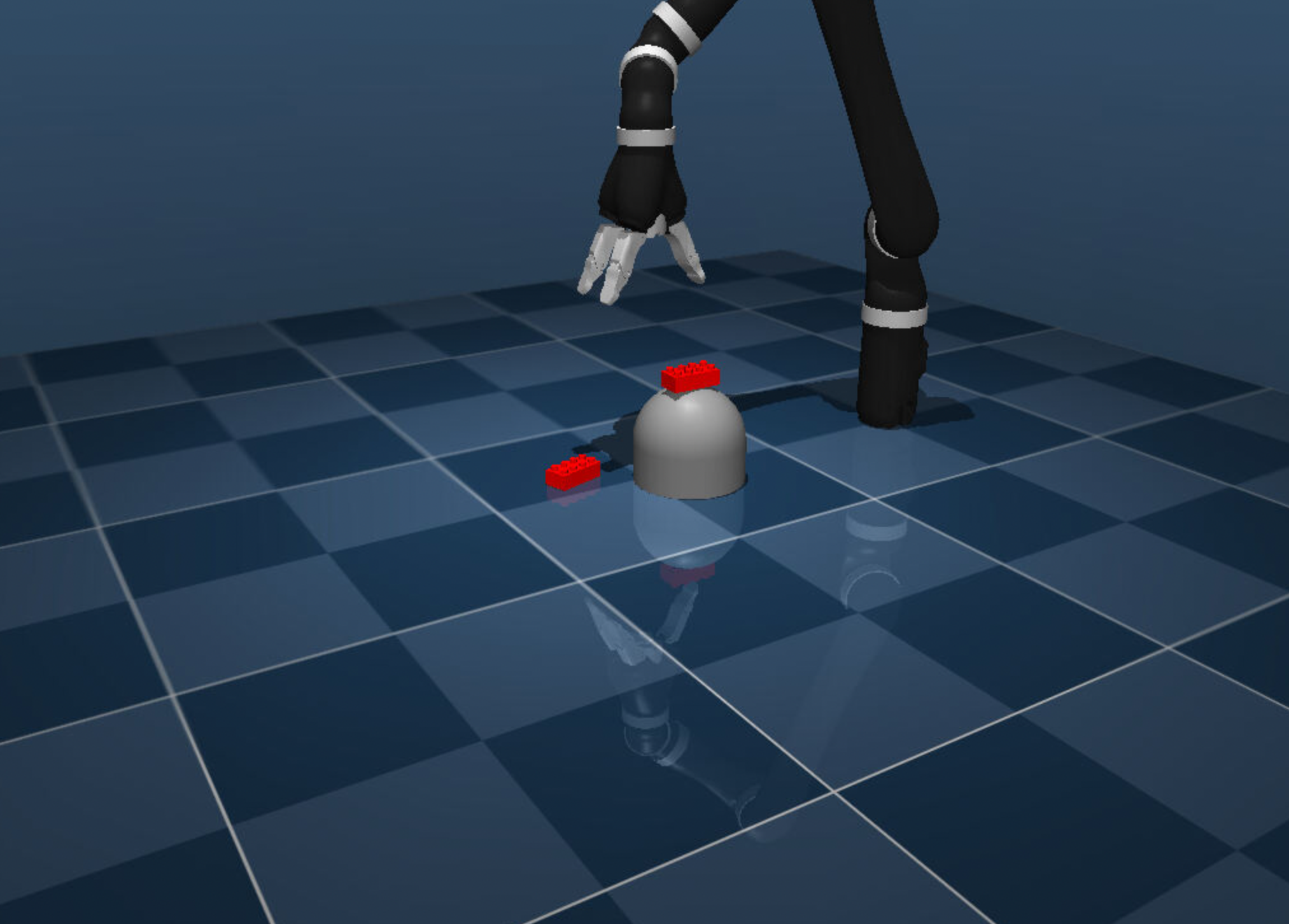}}
	\setcounter {subfigure} {0} c){\includegraphics[width=105pt, height=90pt]{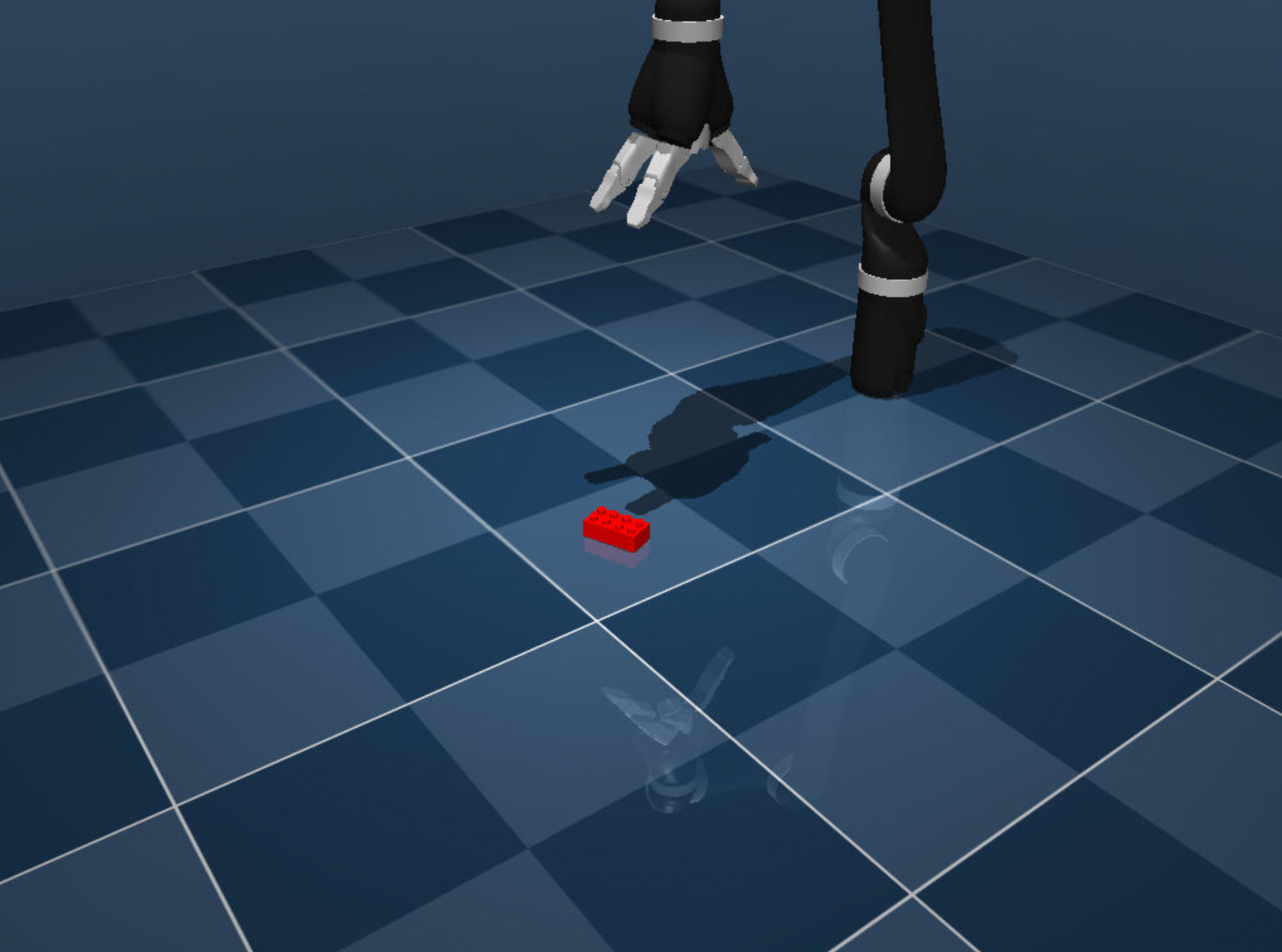}}
 	\setcounter {subfigure} {0} d){\includegraphics[width=105pt, height=90pt]{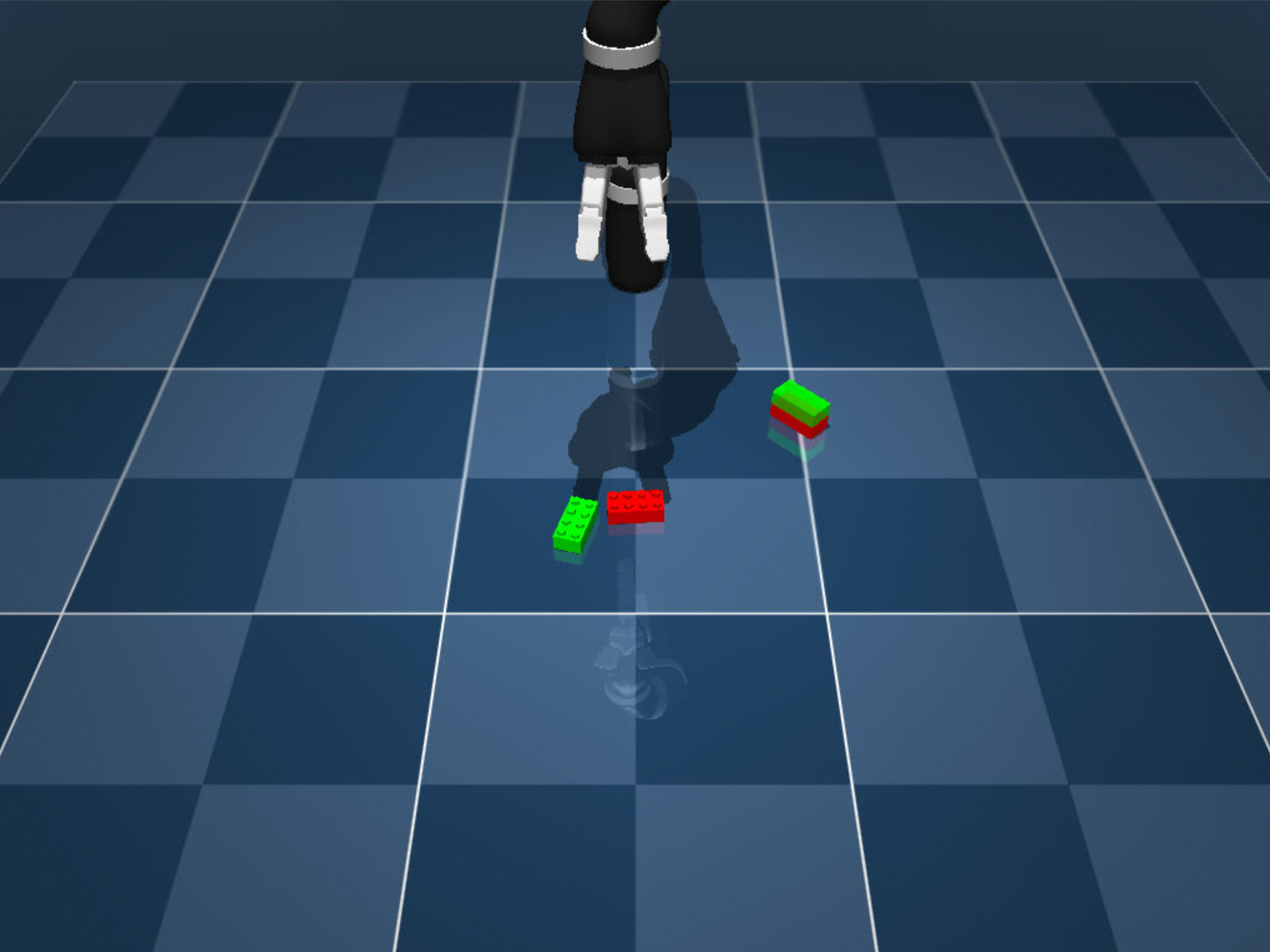}}
	\\
	\setcounter {subfigure} {0} e){\includegraphics[width=105pt, height=90pt]{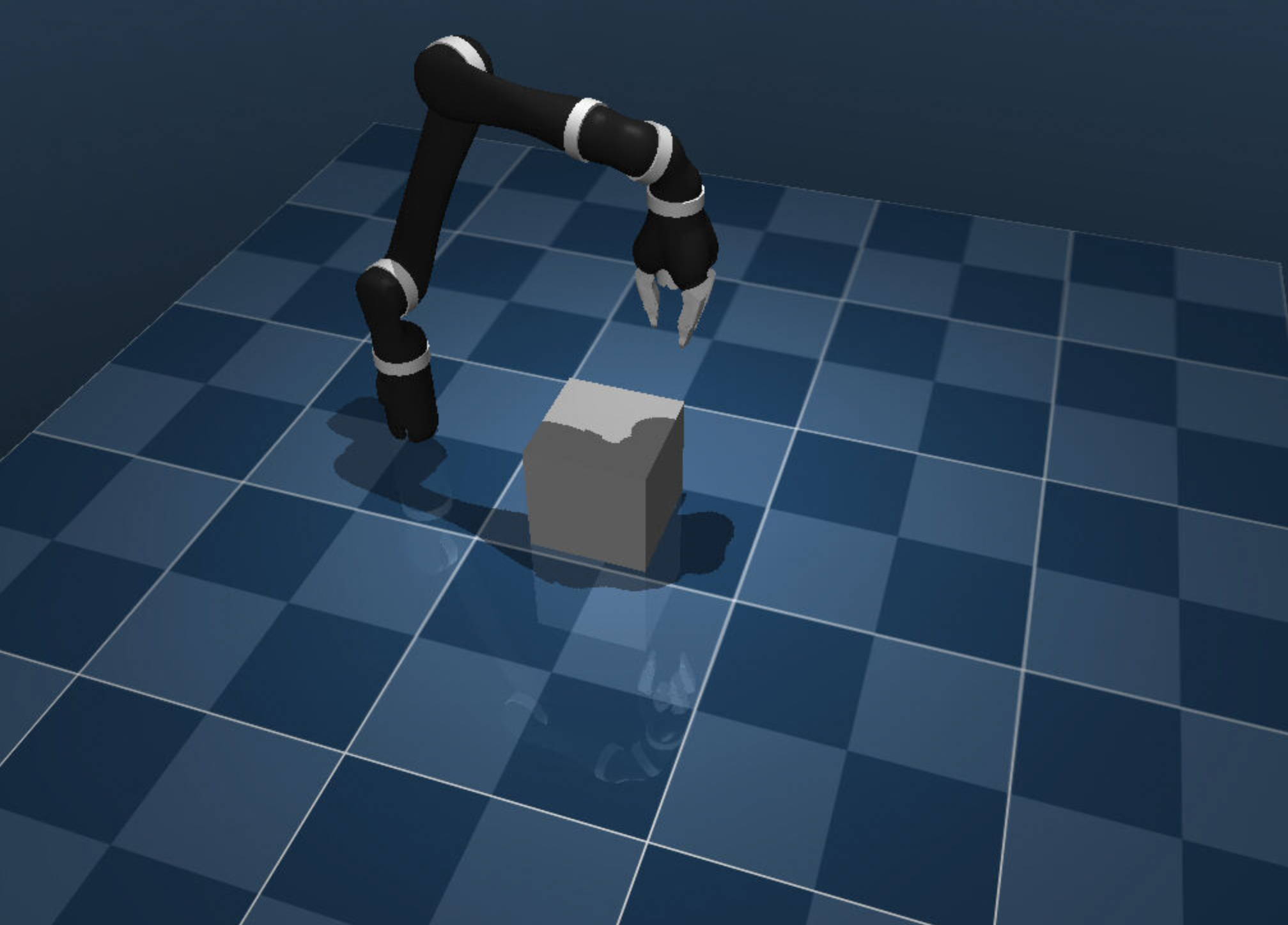}}
	\setcounter {subfigure} {0} f){\includegraphics[width=105pt, height=90pt]{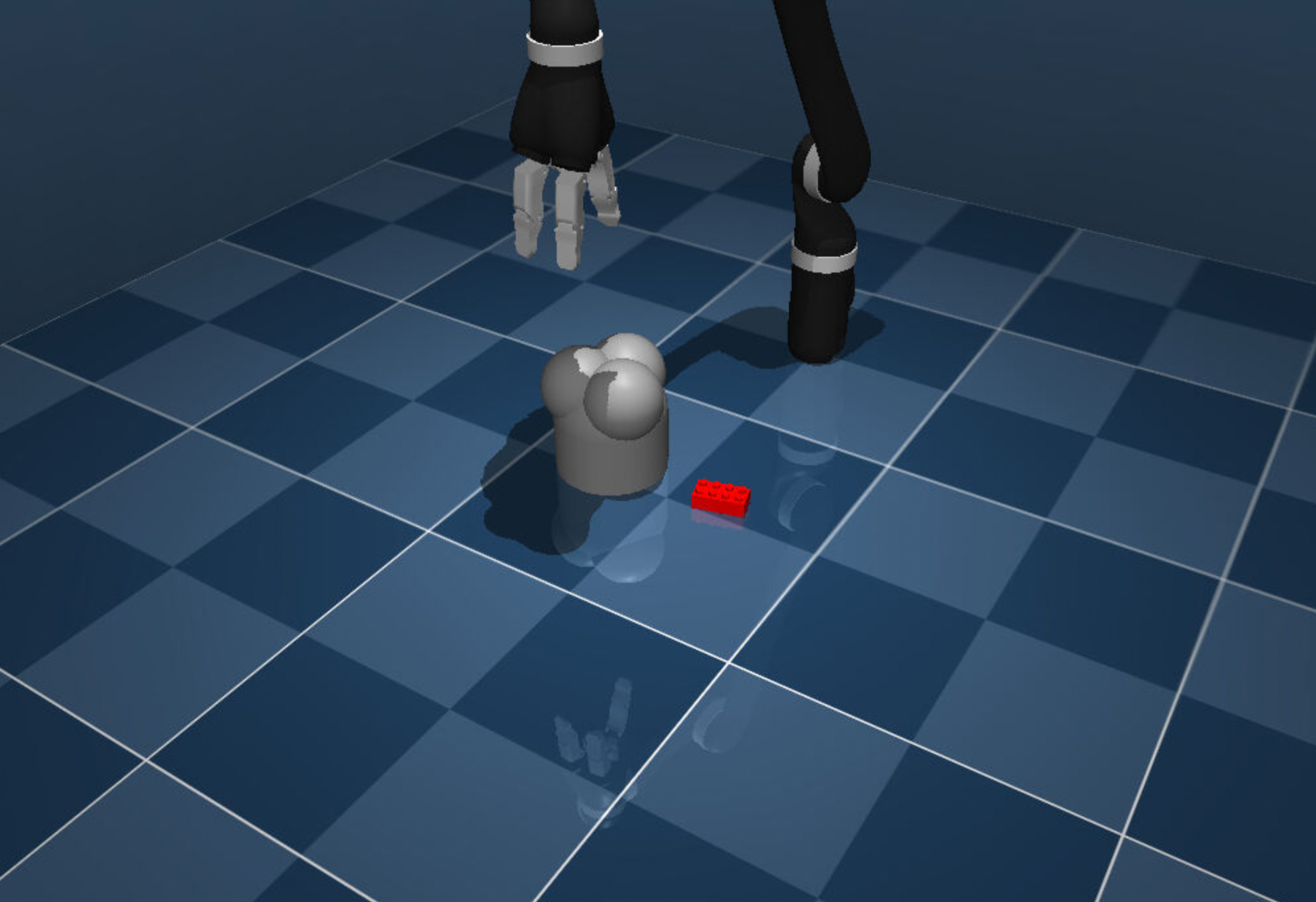}}
	\setcounter {subfigure} {0} g){\includegraphics[width=105pt, height=90pt]{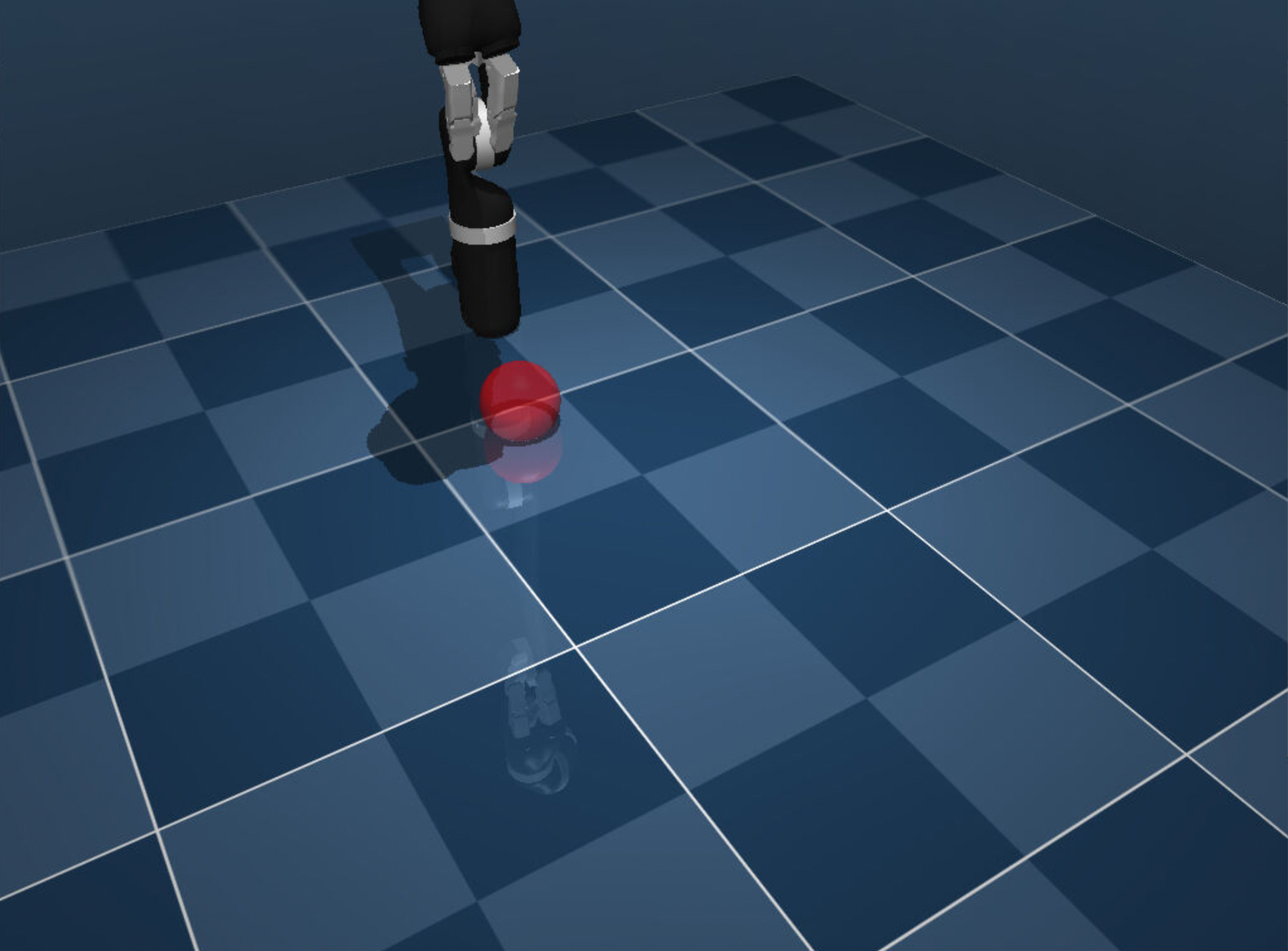}}
 	\setcounter {subfigure} {0} h){\includegraphics[width=105pt, height=90pt]{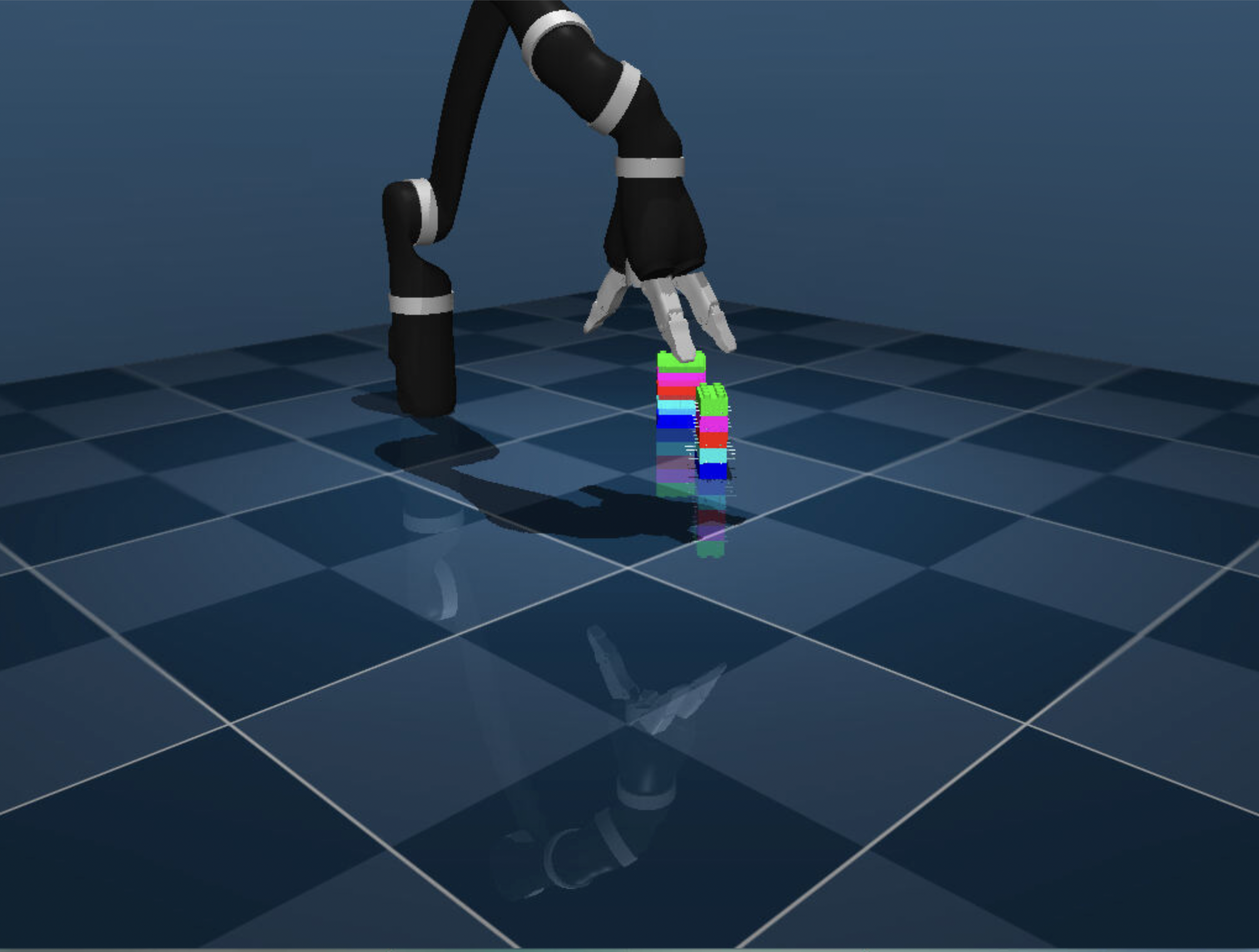}}
  
	\caption{RMBench consists of eight robotic manipulation tasks that offer various challenges, including lifting (a,e), placing (b,f), reaching (c,g), stacking (d) and reassembling (h). (a) lift brick, (b) place brick, (c) reach duplo, (d) stack 2 bricks, (e) lift large box, (f) place cradle, (g) reach site, (h) reassemble 5 bricks random order. Each observed state is a $84 \times 84$ pixel visual image~\cite{c40}}
	\label{fig:Manipulation tasks} 
\end{figure*}

\textit{RMBench} focuses on challenging robotic manipulation tasks to provide a useful performance indicator for current RL algorithms. RMBench contains five types of robotic manipulations: lifting, placing, reaching, stacking, and reassembling. Some types have multiple tasks. In total, nine tasks are implemented in RMBench. They are visualized in Figure~\ref{fig:Manipulation tasks}. Every manipulation environment returns a reward $r\left(s,a\right)\in\left[0,1\right]$ per time-step. The state and action spaces in robotic manipulation tasks are high dimensional and continuous. We select seven RL algorithms according to the published time, type of policy, policy optimization method, and applicability to continuous action spaces. To explore the ability of RL algorithms for high-dimensional state and action spaces without any auxiliary data pre-processing enhancement. The agents are trained with the observed raw pixels as inputs. Classic neural networks, such as Feed-forward neural networks (FFN) and Convolutional neural networks (CNN), are implemented to accelerate the training process. 


{\textbf{Problem Definition.}}
We define the following notation that will be used later. Consider a standard infinite-horizon discounted Markov decision process (MDP)~\cite{c42}, defined by tuple $(\mathcal{S}, \mathcal{A}, \mathcal{P}, r, \rho_0, \gamma, T)$, where $\mathcal{S}$ is a (possibly infinite) set of states, $ \mathcal{A} $ is a set of actions, $ \mathcal{P}: \mathcal{S} \times \mathcal{A} \times \mathcal{S} \rightarrow \mathbb{R}_{> 0}$ a transition function that defines a probability distribution over the next state given the current state and action, $ r : \mathcal{S} \times \mathcal{A} \rightarrow \mathbb{R} $ is a reward function, $\rho_0 : \mathcal{S} \rightarrow \mathbb{R}_{>0}$ is an initial state distribution, $\gamma \in (0, 1]$ is a discount factor, and $T$ is a finite horizon.

A policy is a rule used by an agent to decide what actions to take. In deep RL, we work with parameterized policies that we can improve using an optimization technique to change the agent's behavior. Stochastic policies with parameter $\theta$ are often defined by $a_{t} \sim \pi_\theta \left(\cdot \mid s_{t}\right)$. For deterministic policies, it is usually denoted by $a_{t}=\mu_\theta \left(s_{t}\right)$. Let $J(\pi)$ denote its expected discounted reward: $J(\pi)=\mathbb{E}_{\tau}\left[\sum_{t=0}^T \gamma^t r\left(s_t, a_t\right)\right] $, where $\tau=\left(s_0, a_0, s_1, a_1, \ldots\right)$ represents the whole trajectory, $s_0 \sim \rho_0\left(s_0\right)$, $a_t \sim \pi\left(a_t|s_t\right)$, and $s_{t+1} \sim P\left(s_{t+1}|s_t, a_t\right)$. For whatever choice of return measure and whatever choice of policy, the objective of RL is to train a policy which maximizes the expected return.
We use Generalized Advantage Estimation (GAE)~\cite{c109} for computing the policy gradient. Let $V$ be an approximate value function. Define the temporal difference (TD) residual $\delta_t^V$ of $V$ with discount $\gamma$: $ \delta_t^V=r_t+\gamma V\left(s_{t+1}\right)-V\left(s_t\right)$. The sum of $k$ of these $\delta$ terms is defined as  $\hat{A}_t^{(k)}:=\sum_{l=0}^{k-1} \gamma^l \delta_{t+l}^V$. The generalized advantage estimator $\hat{A}_t^{\mathrm{GAE}(\gamma, \lambda)}$ is defined as the exponentially-weighted average of these $k$-step estimators:
\begin{equation}
    \begin{aligned}
    \hat{A}_t^{\mathrm{GAE}(\gamma, \lambda)} &:=(1-\lambda)\left(\hat{A}_t^{(1)}+\lambda \hat{A}_t^{(2)}+\lambda^2 \hat{A}_t^{(3)}+\ldots\right) \\
    &=\sum_{l=0}^{\infty}(\gamma \lambda)^l \delta_{t+l}^V,
    \end{aligned}
    \label{eq:GAE}
\end{equation}
where $\gamma \in[0,1]$ and $\lambda \in[0,1]$.

\renewcommand{\arraystretch}{1.5} 
\begin{table}[t] \footnotesize
\setlength\tabcolsep{0.8pt}
\centering
\caption{\textit{RMBench} covers seven model-free RL algorithms developed between 2000 and 2021. We list the main characteristics of algorithms, such as policy learning type (PLT), policy type (PT), and optimization (Opt). The following abbreviations are used: DP (deterministic policy), SP (stochastic policy), PO (policy optimization), QL (Q-learning), and DA (data augmentation).}
\label{tab:RL algorithms}
\begin{tabular}{ccccccccc}
    \toprule
    \multirow{2}{*}{\textbf{Algorithms}} & \multirow{2}{*}{\textbf{Authors, Date}} & \multicolumn{2}{c}{\textbf{PLT}} & \multicolumn{2}{c}{\textbf{PT}} & \multicolumn{2}{c}{\textbf{Opt}} & \multirow{2}{*}{\textbf{DA}} \cr
    \cmidrule(lr){3-4} \cmidrule(lr){5-6} \cmidrule(lr){7-8} 
    & & \textbf{on-policy} & \textbf{off-policy} & \textbf{DP} & \textbf{SP} & \textbf{PO} & \textbf{QL} & \cr
    \midrule
    VPG~\cite{c107}	& Sutton et al., 2000 & $\checkmark$ & $\times$ & $\times$ & $\checkmark$ & $\checkmark$ & $\times$ & $\times$ \cr
    TRPO~\cite{c110} & Schulman et al., 2015 & $\checkmark$ & $\times$ & $\times$ & $\checkmark$ & $\checkmark$ & $\times$ & $\times$ \cr
    PPO~\cite{c111}	& Schulman et al., 2017 & $\checkmark$ & $\times$ & $\times$ & $\checkmark$ & $\checkmark$ & $\times$ & $\times$ \cr    
    DDPG~\cite{c112} & Silver et al., 2014  & $\times$ & $\checkmark$ & $\checkmark$ & $\times$ & $\checkmark$ & $\checkmark$ & $\times$ \cr
    TD3~\cite{c114}	& Fujimoto et al., 2018  & $\times$ & $\checkmark$ & $\checkmark$ & $\times$ & $\checkmark$ & $\checkmark$ & $\times$ \cr
    SAC~\cite{c115}	& Haarnoja et al., 2018  & $\times$ & $\checkmark$ & $\times$ & $\checkmark$ & $\checkmark$ & $\checkmark$ & $\times$ \cr
    DrQ-v2~\cite{c103}  & Yarats et al., 2021 & $\times$ & $\checkmark$ & $\checkmark$ & $\times$ & $\checkmark$ & $\checkmark$ & $\checkmark$ \cr
    \bottomrule
\end{tabular}
\end{table}

\textbf{Benchmarked Algorithms.}
We select the representative algorithms in RMBench based on the published time (from 2000 to 2021), policy (stochastic or deterministic), policy learning strategy (on-policy or off-policy), policy optimization method (policy optimization, Q-learning, or combination of both), and applicability to continuous action spaces. Their main characteristics and differences are compared in Table~\ref{tab:RL algorithms}. 

\begin{itemize}[itemsep=0pt,topsep=0pt,parsep=0pt,leftmargin=*]
    \item \textit{Vanilla Policy Gradient (VPG)} \cite{c107}\cite{c108}. VPG optimizes the stochastic policy directly by gradient ascent to maximize performance. The optimization is performed on-policy, which means that each update uses experiences collected by the most recent version of the policy. It provides a nice reformation of the derivative of the objective function and simplifies gradient computation a lot. Since the value of the infinite number of actions and states needed to be estimated, this policy-based algorithm is useful in the continuous space.
    
    \item \textit{Trust Region Policy Optimization (TRPO)}~\cite{c110}. To improve training monotonically, TRPO updates policies when satisfying a special constraint. The constraint is KL-Divergence, a measure of distance between probability distributions. Benefiting from the constraint, TRPO enforces the distance between old and new policies to be small enough. Compared with VPG, TRPO nicely avoids the collapse when using large step sizes to update policies’ parameters. Therefore, TRPO can guarantee a stable improvement over policy iteration.

    \item \textit{Proximal Policy Optimization (PPO)}\cite{c111}. PPO is a family of first-order methods for keeping new policies close to old, focusing on the performance collapse caused by large updating steps. While TRPO attempts to solve this issue using a complicated second-order method. PPO simplifies TRPO by employing a clipped surrogate objective without KL-divergence while preserving performance. PPO-Clip is a variant of PPO. It imposes the constraint by squeezing the distance between the new and old policies into a small, predetermined interval.

    \item \textit{Deep Deterministic Policy Gradient (DDPG)}~\cite{c112}. DDPG is an off-policy algorithm that updates a deterministic policy, unlike the aforementioned algorithms, such as VPG, TRPO, and PPO, which train a stochastic policy in an on-policy way. It combines Deterministic Policy Gradient (DPG) ~\cite{c113} and Deep Q-Network (DQN) ~\cite{c100} and simultaneously learns a Q-function and a policy in the actor-critic framework. To establish an exploration policy, noise is added to DDPG policies to make them explore more.

    \item \textit{Twin Delayed DDPG (TD3)}\cite{c114}. TD3 applied three key tricks to DDPG to prevent the overestimation of the value function, given that DDPG typically suffers from an overestimation of the value function. Clipped Double Q-learning was developed to favor underestimation bias. The Delayed update of the Target and Policy Networks strategy was implemented to reduce estimation variance and stabilize the training process. And TD3 employed a smoothing regularization strategy on the value function to avoid the estimation of local peaks in value.

    \item \textit{Soft Actor-Critic (SAC)}~\cite{c115}. SAC trains a stochastic policy with entropy regularization and explores in an on-policy way.  SAC is an actor-critic algorithm based on the entropy-regularized RL framework ~\cite{c118}, which optimizes policy and value function networks separately. Unlike TD3, SAC trains the policy with the objective to maximize the expected return and the entropy simultaneously. The Entropy maximization leads to more exploration. Like TD3, SAC uses the clipped double-Q trick and takes the minimum Q-value between the two Q approximators. In our benchmark, the entropy regularization coefficient is fixed.

    \item \textit{DrQ-v2}\cite{c103}. Since data augmentations are essential to several recent visual RL algorithms~\cite{c121, c122, c123}.To solve tasks directly from pixel observations, DrQ-v2 applies the image augmentation of random shifts to pixel observations of states. It uses DDPG~\cite{c112} as a backbone actor-critic RL algorithm with $n$-step returns~\cite{c117}, resulting in faster reward propagation and overall learning progress~\cite{c119}. Additionally, the exploration noise variance is handled by linear decay of the exploration schedule ~\cite{c124}
\end{itemize}

\definecolor{color_best}{RGB}{22, 138, 173}

\begin{table*}[t]
\vspace{4mm}
\caption{The MER performance (mean and the standard deviation) of the seven RL algorithms with the nine robotic manipulation tasks, which is averaged over 200 episodes and five different random seeds. The best, second, and third best results are highlighted by different shades of blue, with the darkest blue indicating the best.}
\label{table:Performance results}
\centering
\resizebox{\textwidth}{!}{%
\begin{tabular}{c|ccccccc}
\hline
Task & VPG & TRPO & PPO & DDPG & TD3 & SAC & DrQ-v2 \\

\hline
Place brick & $4.360\pm0.085$ & \cellcolor{color_best!15}{$4.723\pm0.162$} & $4.371\pm0.090$ & $4.314\pm0.096$ & $4.372\pm0.086$ & \cellcolor{color_best!60}{$5.03\pm0.058$} & \cellcolor{color_best!30}{$4.429\pm0.101$} \\

Place cradle & $4.319\pm0.059$ & \cellcolor{color_best!30}{$4.552\pm0.121$} & \cellcolor{color_best!15}{$4.385\pm0.058$} & $4.305\pm0.079$ & $4.339\pm0.17$ & \cellcolor{color_best!60}{$4.942\pm0.102$} & $4.308\pm0.178$ \\

\hline
Reach duplo & \cellcolor{color_best!15}{$2.832\pm1.778$} & \cellcolor{color_best!30}{$7.296\pm1.465$} & $2.399\pm1.501$ & $0.589\pm0.553$ & $0.923\pm1.765$ & \cellcolor{color_best!60}{$9.297\pm0.635$} & $1.906\pm1.135$ \\

Reach site & $1.217\pm0.26$ & \cellcolor{color_best!30}{$3.975\pm0.719$} & $1.161\pm0.270$ & $1.380\pm0.090$ & $1.218\pm0.148$ & \cellcolor{color_best!60}{$4.519\pm0.172$} & \cellcolor{color_best!15}{$1.904\pm0.770$} \\

\hline
Lift large box & \cellcolor{color_best!30}{$0.003\pm0.001$} & \cellcolor{color_best!15}{$0.002\pm0.000$} & \cellcolor{color_best!30}{$0.003\pm0.001$} & \cellcolor{color_best!30}{$0.003\pm0.001$} & \cellcolor{color_best!60}{$0.004\pm0.002$} & \cellcolor{color_best!15}{$0.002\pm0.000$} & \cellcolor{color_best!15}{$0.002\pm0.000$} \\

Lift brick &  \cellcolor{color_best!60}{$0.002\pm0.000$} & \cellcolor{color_best!35}{$0.001\pm0.000$} & \cellcolor{color_best!35}{$0.001\pm0.000$} & \cellcolor{color_best!35}{$0.001\pm0.000$} & \cellcolor{color_best!60}{$0.002\pm0.001$} & \cellcolor{color_best!35}{$0.001\pm0.000$} & \cellcolor{color_best!35}{$0.001\pm0.001$} \\

\hline   
Stack 2 breaks & \cellcolor{color_best!60}{$0.148\pm0.011$} & $0.144\pm0.013$ & \cellcolor{color_best!15}{$0.145\pm0.013$} & \cellcolor{color_best!60}{$0.148\pm0.011$} & \cellcolor{color_best!15}{$0.145\pm0.012$} & \cellcolor{color_best!30}{$0.147\pm0.013$} & $0.142\pm0.008$ \\

Stack 2 breaks movable base & \cellcolor{color_best!35}{$0.147\pm0.009$} & \cellcolor{color_best!10}{$0.144\pm0.013$} & $0.143\pm0.009$ & \cellcolor{color_best!60}{$0.150\pm0.008$} & \cellcolor{color_best!30}{$0.147\pm0.012$} & $0.143\pm0.013$ & $0.142\pm0.018$ \\

\hline
Reassemble 5 bricks random order & \cellcolor{color_best!30}{$60.543\pm0.779$} & $59.893\pm1.590$ & \cellcolor{color_best!15}{$60.347\pm2.280$} & $60.244\pm1.712$ & $59.018\pm2.068$ & $56.769\pm1.180$ & \cellcolor{color_best!60}{$61.131\pm2.912$} \\

\bottomrule
\end{tabular}
}
\end{table*}
\vspace{-2mm}

\begin{figure*}
    \centering
    \includegraphics[width=\textwidth]{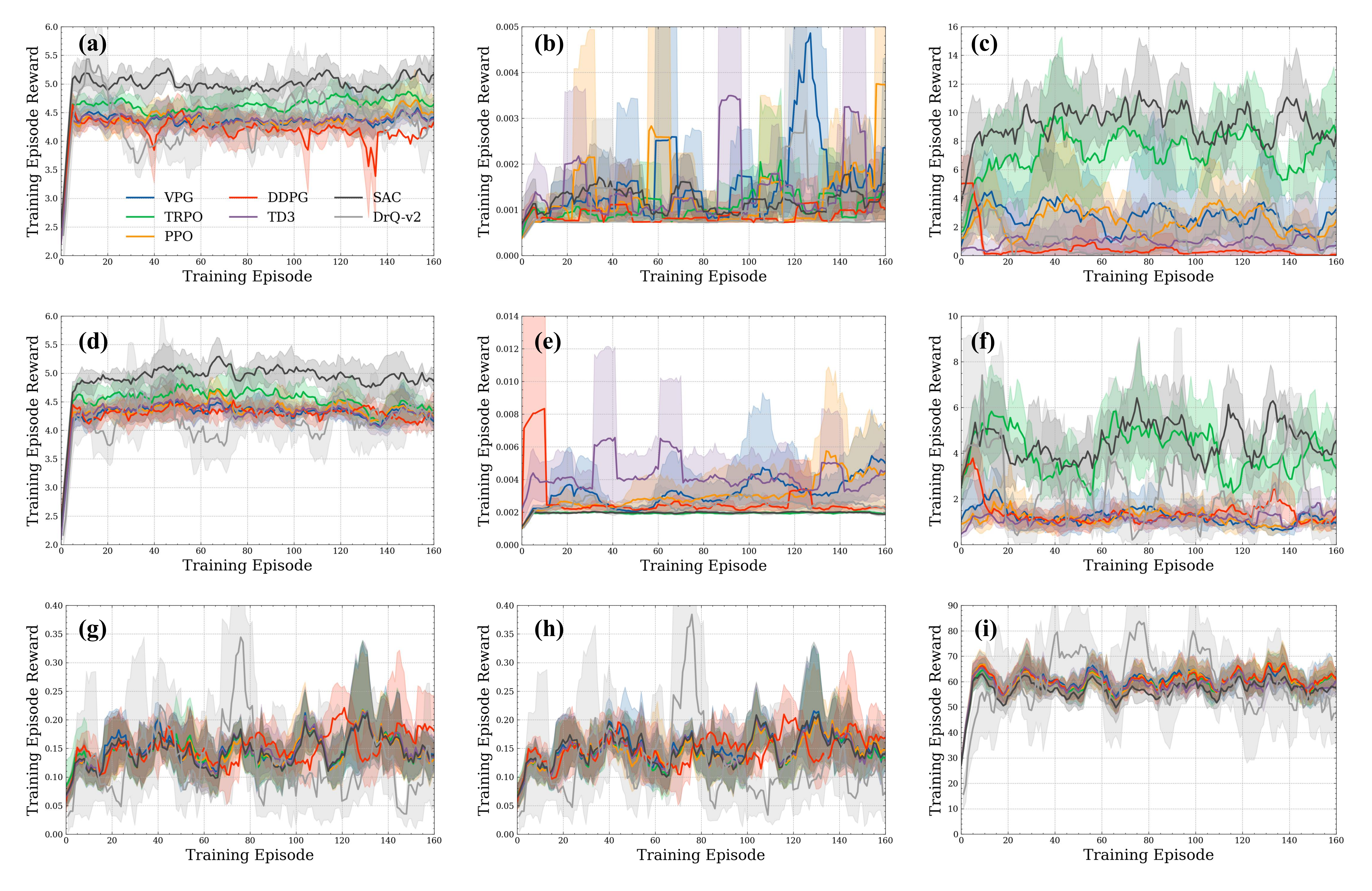}
    \vspace{-2mm}
    \caption{Training convergence curves of the seven algorithms on the nine robotic manipulation tasks: (a) place brick, (b) lift brick, (c) reach duplo, (d) place cradle, (e) lift large box, (f) reach site, (g) stack 2 bricks, (h) stack 2 bricks movable base, (i) reassemble 5 bricks random order. Shaded regions correspond to one standard deviation. }
    \label{fig:Convergence curves}
\end{figure*}

\section{Experiments}

\subsection{Experimental Settings}

\smallskip
\noindent \textbf{Setup}. We utilize \textit{dm\_control} software package~\cite{c40}, which has task suites for reinforcement learning agents in an articulated-body simulation. It relies on Multi-Joint dynamics with Contact (MuJoCo)~\cite{c31}, a fast and accurate physics simulator~\cite{c41}. We focus on the manipulation tasks with a 3D robotic arm, such as lifting, placing, reaching, stacking, and reassembling. The policies are and trailearnedning policies from observed raw statespixels directly. Each observed state is a visual image of $84\ \times84$ pixels. All of the manipulation environments return a reward $r(s,a) \in [0, 1]$ per timestep and have a time limit of 10 seconds.

\smallskip
\noindent \textbf{Implementation Details}. For all tasks, we train the agent for 200 episodes with five different random seeds. An episode includes 2000 steps.  For each algorithm's implementation, we refer to the existing public code\footnote{\scriptsize \url{https://github.com/openai/spinningup} and \url{https://github.com/facebookresearch/drqv2}} with moderate modifications. We run experiments with the same training configuration and the default values of hyper-parameters as described in its original paper. Since thea vanilla policy gradient update  has no bias but high variance, we utilize an effective variance reduction scheme for policy gradients: General Advantage Estimation (GAE)~\cite{c109}, see Eq.~(\ref{eq:GAE}), for on-policy algorithms, such as VPG, TRPO, and PPO, see Eq.~(\ref{eq:GAE}). The $\lambda$ in GAE is set to 0.95. During training for actor-critic algorithms such as DDPG, TD3, and SAC, we leverage a number of techniques to improve exploration and stability. Before running the real policy, we set a few steps (10000) for random action selection. Second, gradient descent updates until 1000 interactions to verify that the replay buffer is sufficiently full to support useful updates. Thirdly, we update the networks every 50 environment interactions, meaning that the actor and critic networks are bound to change slowly, and the learning stability is significantly improved. And we add the Gaussian exploration noise (standard deviation=0.1) to policy at training time.

\smallskip
\noindent \textbf{Network Architecture}.
For all algorithms (excluding DrQ-v2), we use the same feed-forward neural network policy with three hidden layers, consisting of 512, 256, and 128 hidden units with tanh activation at the three hidden layers. Their agents are trained based on raw flattened pixels in RGB. For the DrQ-v2 algorithm, following its original setting, we use four convolution layers (without pooling) with 32 filters at each layer, followed by two fully connected layers with 1024 units. The inputs of DrQ-v2 are the image augmentation of random shifts to pixel observations of states.
\vspace{-1em minus 0.5em}

\subsection{Performance Metrics}
\vspace{0em minus 0.8em}

To measure the performance of an episode during training, we use \emph{Episode Reward} (\emph{ER}):
\begin{equation}
    ER^i=\sum_{t=1}^{T}R_t^i
    \label{eq:Episode Reward(ER)}
\end{equation}
where $T$ is the number of steps in an episode, and $R_t^i$ is the undiscounted return for the $t$-th step of the $i$-th episode. The average performance for a specific task is measured by the \emph{Mean Episode Reward} (\emph{MER}) : 
\begin{equation}
    MER=\frac{1}{I}\sum_{i=1}^{I}{ER^i}=\frac{1}{I}\sum_{i=1}^{I}\sum_{t=1}^{T}R_t^i
    \label{eq:Mean Episode Reward(MER)}
\end{equation}
where $I$ is the number of episodes.

\begin{figure}[h]
    \centering
    \includegraphics[width=0.45\textwidth]{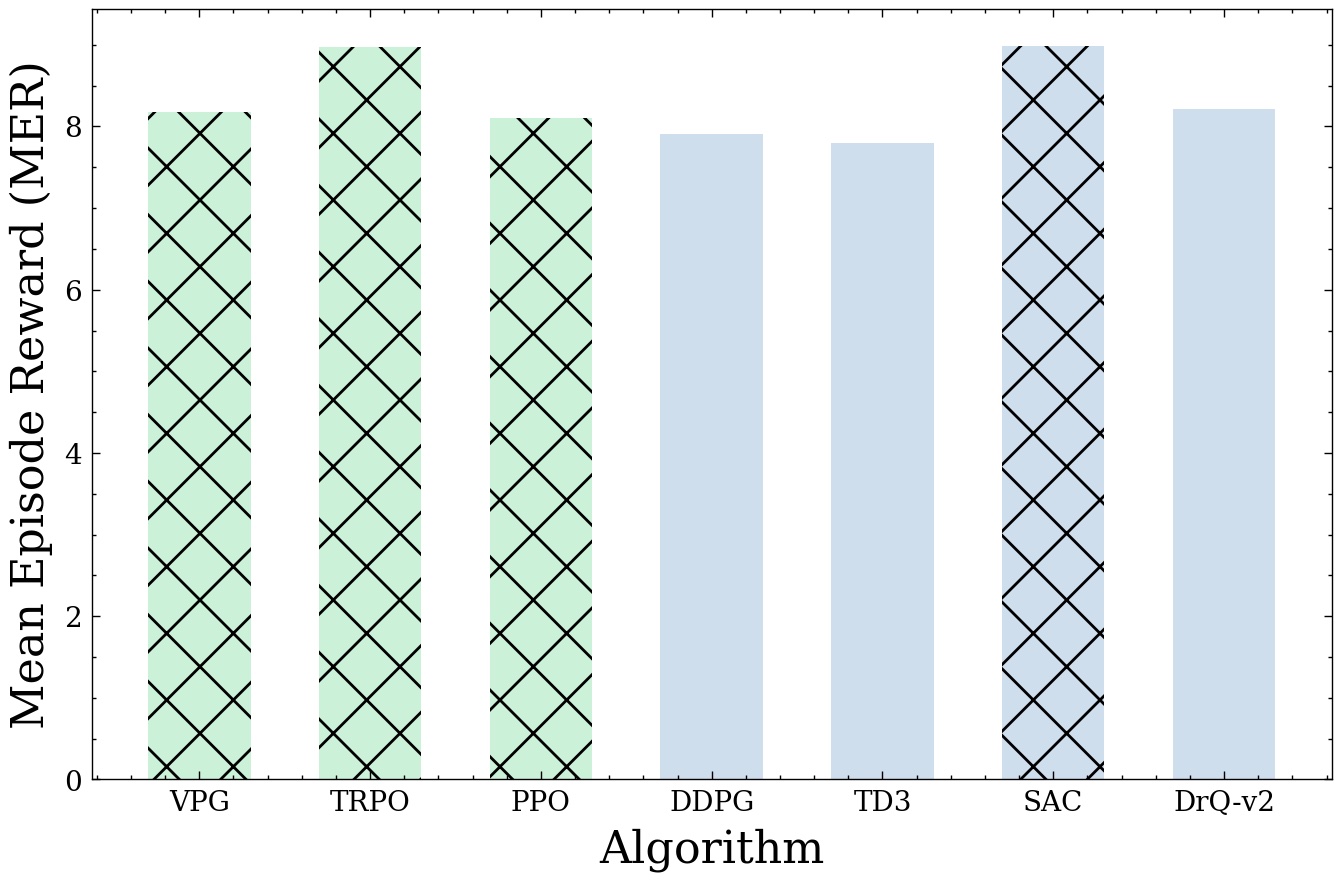}
    \caption{The average performance of each algorithm for all the benchmark tasks in \textit{RMBench}. The $y$-axis demonstrates Mean Episode Reward (MER). The $x$-axis means RL algorithms. Green bars mean algorithms using policy optimization. Blue bars mean algorithms using the actor-critic mechanism. Bars filled with slanted lines indicate algorithm using stochastic policy, and bars not filled indicate algorithm using deterministic policy.}
    \label{fig:algo_bar}
\end{figure}

\subsection{Results and Discussion}

The overall results are displayed in Table~\ref{table:Performance results},  Fig.~\ref{fig:Convergence curves}, and Fig.~\ref{fig:algo_bar}. Specifically, the average performance of all episodes is reported in Table~\ref{table:Performance results}, where a pair of numbers in each entry represents the MER's mean and the standard deviation. Fig.~\ref{fig:Convergence curves} depicts the training convergence curves for the five initialization seeds at each episode. The $x$-axis shows the training episode. The $y$-axis shows the Episode Reward (ER). Solid lines are average values over five random runs. {Note that we smooth the curve by averaging across ten episodes like in~\cite{c22, c44}}. Shaded regions correspond to one standard deviation. Furthermore, we display the average performance of each algorithm for all tasks in Fig.~\ref{fig:algo_bar}. From these results, we have the following observations:


\noindent \textbf{Algorithm Level.}
1) \underline{\textit{VPG}}: Even though VPG is a simple algorithm, its performance is acceptable for most robotic manipulation tasks. However, VPG sometimes converges too rapidly to a locally optimal, as shown in Fig.~\ref{fig:Convergence curves}(a)(c)(d)(f), which is also pointed out by Peters et al.~\cite{c26}.
2) \underline{\textit{TRPO}}: TRPO shows competitive results on robotic manipulation tasks, such as placing and reaching, suggesting that update the policy when satisfying the KL-Divergence constraint is beneficial to ensuring consistent improvement over policy iteration.
3) \underline{\textit{PPO}}: PPO can be viewed as a simplified version of TRPO. Our simulations show that PPO achieves worse performance than TRPO. One reason is that the clipped-PPO is unstable when rewards vanish outside bounded support on continuous action spaces. And it is sensitive to policy initialization when there are locally optimal actions close to the initialization~\cite{c27}.
4) \underline{\textit{DDPG}}: The performance of DDPG is mediocre. Since DDPG is an off-policy algorithm, large replay buffers allow it to benefit from learning over a collection of uncorrelated transitions~\cite{c112}. Furthermore, it is critical to choose appropriate noises that facilitate temporally correlated exploration in the physical control problem~\cite{c45}.
5) \underline{\textit{TD3}}: Based on DDPG, TD3 applies a couple of tricks, such as the Clipped Double Q-learning and “Delayed” Policy Updates. However, the simulation results show that the performance of TD3 is similar to that of DDPG, which indicates that these tricks cannot boost performance significantly in these tasks.
6) \underline{\textit{SAC}}: Regarding average reward and stability, SAC shows the best performance. A central feature of SAC is the entropy regularization. The simulation suggests that the entropy regularization has a better trade-off between exploration and exploitation, which encourages greater exploration and keeps the policy from prematurely converging to an inferior local optimum.
7) \underline{\textit{DrQ-v2}}: DrQ-v2 can be regarded as adding data augmentation and the convolutional encoder to the DDPG.  In terms of MER, DrQ-v2 performs better than DDPG. This improvement should be due to data augmentation and the convolutional encoder, which can help the agent obtain richer environmental information and enables the agent to learn policies more efficiently. However, the training phase is unstable according to the standard deviations and fluctuant training curves. The reason could be that the state details are lost due to convolution operations.

\noindent \textbf{ Category Level.}
1) \underline{\textit{On-policy and off-policy}}:
Our benchmark uses three on-policy algorithms (VPG, TRPO, and PPO) and four off-policy algorithms (DDPG, TD3, SAC, and DrQ-v2). From Fig. \ref{fig:algo_bar}, we can find that the performances of the two types of algorithms are similar. Furthermore, TRPO is the most potent on-policy algorithm, and SAC is the best off-policy algorithm. SAC is marginally better than TRPO, which implies that the sample collection follows the off-policy way bringing better exploration.

2) \underline{\textit{Only policy optimization and actor-critic}}: The major strength of policy optimization is that we can optimize policy directly, which makes algorithms more principled and stable. In addition, Q-learning optimizes policy indirectly to improve performance, which tends to be less stable but better sample efficiency. Actor critic methods live in between policy optimization and Q-learning. From Fig.~\ref{fig:algo_bar}, the performance of actor-critic algorithms such as DDPG, TD3, SAC, and DrQ-v2 achieve comparable performance to policy optimization methods, which suggests that actor-critic approaches can provide a satisfactory compromise between policy optimization and Q-Learning.

3) \underline{\textit{Deterministic policy and stochastic policy}}: In order to explore the whole state and action spaces, a stochastic policy is often applicable, in which the policy gradient should be computed by integrating both state and action spaces. Whereas deterministic policy only integrates over the state space. For more exploration, it often adds noise to deterministic policy. The deterministic policy provides a significant performance benefit for high-dimensional tasks, as pointed out by Silver et al.~\cite{c113}. From Fig.~\ref{fig:algo_bar}, the performance of algorithms based on the stochastic policy such as VPG, TRPO, PPO,  and SAC achieve comparable performance to algorithms based on the deterministic policy. As a result, the stochastic policy is more capable of producing satisfactory results, which implies the importance of exploration in the learning phase. 

3) \underline{\textit{Data Augmentation}}: DrQ-v2 is an RL algorithm based on DDPG within data augmentation and convolutional encoder. The simulations show that DrQ-v2 outperforms DDPG in most tasks, such as placing, reaching and reassembling. And DrQ-v2 achieves the best performance in reassembling 5 bricks in random order task. Therefore, the data augmentation and convolutional encoder can facilitate agents to understand the pixel-level states. Since that the data augmentation strategy can be easily exploited by all RL algorithms in visual control tasks. A combination of advanced RL algorithms, such as SAC, and data augmentation can achieve more satisfactory results. Specifically, a larger replay buffer is crucial, and this becomes more apparent in tasks with a wider variety of beginning state distributions ~\cite{c46}. Therefore, a substantial replay buffer is necessary when using data augmentation.

\noindent \textbf{Task Level.}
Different algorithms produce different results for various tasks. SAC and TRPO perform better than other algorithms for placing manipulation, including brick and cradle. SAC is the most effective in reaching tasks, including duplo and reaching site, followed by TRPO. DrQ-v2 obtains the best mean episode reward in reassembling five bricks.  We find that all the algorithms have disappointing performance on the lifting and stacking tasks from the unstable training curves and mean episode rewards. The reason can be that these tasks require more precise and complex arm and finger joint coordination.

\section{Conclusion and Future Work}
In this paper, we present \textit{RMBench}, the first benchmark for robot manipulations to evaluate the human-like abilities of seven RL algorithms by directly learning dexterous manipulations from observed raw pixels. \textit{RMBench} contains nine robotic manipulation tasks that cover lifting, placing, reaching, stacking, and reassembling. We implement and evaluate seven model-free RL algorithms. We provide a standardized set of RL-based benchmarking experiments as baselines for visual robotic manipulation tasks. Our simulations show that none of the algorithms can handle all tasks well. Although VPG is the simplest algorithm, it performs satisfactorily in most benchmark tasks. SAC is an effective method for training deep neural network policies regarding average reward and stability among the tested algorithms. Data augmentation and autoencoder used in DrQ-v2 can help agents gain richer information about environments, even though the learning of DrQ-v2 is unstable. The implemented RL algorithms all show poor performance on the lifting tasks, which calls for developing new and better algorithms. Since that agent trained in the 3D simulation environment can be transferred to real-world tasks. The results provide a guideline for leveraging RL algorithms in real-world robotic manipulation tasks. Overall, the goal of RMBench is to facilitate the study and improvement of robot manipulation solutions by providing a simulation framework and environment benchmark.


\textbf{Future Works}. 
More robotic manipulation tasks should be involved in RMBench to enrich the diversity of tasks.  Since that there are inherent discrepancies between the dynamics of a simulator and the real world. Future research should evaluate whether simulator-trained agents can be transferred to the real-world situations effortlessly. Given that the robotic simulation environments are more complex than the 2D game due to the 3D distances and the variations of shadow and light intensity. The strategy to enhance understanding and extraction of state information should be considered. The deeper neural networks should be applied to enhance the understanding of states.  And various data augmentation methods leveraged in the Computer Vision field could be considered to improve the sample efficiency in the learning phase. Furthermore, the time cost of the different algorithms needs to be compared in the future. It is necessary to fix the configuration of the CPU and GPU to record the wall-clock time in training. We leave it for future work.


\begin{thebibliography}{99}

\bibitem{c100} V. Mnih et al., “Human-level control through deep reinforcement learning,” Nature, vol. 518, no. 7540, Art. no. 7540, Feb. 2015, doi: 10.1038/nature14236.

\bibitem{c101} X. Guo et al., “Deep Learning for Real-Time Atari Game Play Using Offline Monte-Carlo Tree Search Planning,” NIPS, 2014, vol. 27. 

\bibitem{c102} M. Watter et al., “Embed to Control: A Locally Linear Latent Dynamics Model for Control from Raw Images,” NIPS, 2015, vol. 28.

\bibitem{c103} D. Yarats et al., “Mastering Visual Continuous Control: Improved Data-Augmented Reinforcement Learning,” arXiv:2107.09645 [cs], Jul. 2021.

\bibitem{c104} M. G. Bellemare et al., “The Arcade Learning Environment: An Evaluation Platform for General Agents,” JAIR, vol. 47, pp. 253–279, Jun. 2013, doi: 10.1613/jair.3912.

\bibitem{c105} Y. Duan et al., “Benchmarking Deep Reinforcement Learning for Continuous Control,” ICML, Jun. 2016, pp. 1329–1338.

\bibitem{c106} H. Nguyen and H. La, “Review of Deep Reinforcement Learning for Robot Manipulation,” IRC, Feb. 2019, pp. 590–595.

\bibitem{c107} R. S. Sutton et al., “Policy Gradient Methods for Reinforcement Learning with Function Approximation,” NIPS, 1999, vol. 12.

\bibitem{c108} J. Schulman, “Optimizing Expectations: From Deep Reinforcement Learning to Stochastic Computation Graphs,” UC Berkeley, 2016.

\bibitem{c109} J. Schulman et al., “High-Dimensional Continuous Control Using Generalized Advantage Estimation,” arXiv, arXiv:1506.02438, Oct. 2018.

\bibitem{c110} J. Schulman et al., “Trust Region Policy Optimization,” ICML, Jun. 2015, pp. 1889–1897.

\bibitem{c111} J. Schulman et al., “Proximal Policy Optimization Algorithms.” arXiv, Aug. 28, 2017.

\bibitem{c112} T. P. Lillicrap et al., “Continuous control with deep reinforcement learning,” arXiv, arXiv:1509.02971, Jul. 2019.

\bibitem{c113} D. Silver et al., “Deterministic Policy Gradient Algorithms,” ICML, Jan. 2014, pp. 387–395.

\bibitem{c114} S. Fujimoto et al., “Addressing Function Approximation Error in Actor-Critic Methods,” arXiv, arXiv:1802.09477, Oct. 2018.

\bibitem{c115} T. Haarnoja et al., “Soft Actor-Critic: Off-Policy Maximum Entropy Deep Reinforcement Learning with a Stochastic Actor,” arXiv:1801.01290 [cs, stat], Aug. 2018.

\bibitem{c116} T. Haarnoja et al., “Reinforcement Learning with Deep Energy-Based Policies,” arXiv:1702.08165 [cs], Jul. 2017. 

\bibitem{c117} G. Barth-Maron et al., “Distributed Distributional Deterministic Policy Gradients,” arXiv, arXiv:1804.08617, Apr. 2018.

\bibitem{c118} B. D. Ziebart et al., “Maximum entropy inverse reinforcement learning.,” in Aaai, 2008, vol. 8, pp. 1433–1438.

\bibitem{c119} V. Mnih et al., “Asynchronous Methods for Deep Reinforcement Learning,” ICML, Jun. 2016, pp. 1928–1937. 

\bibitem{c121} A. Srinivas et al., “CURL: Contrastive Unsupervised Representations for Reinforcement Learning,” arXiv, arXiv:2004.04136, Sep. 2020.

\bibitem{c122} R. Raileanu et al., “Automatic Data Augmentation for Generalization in Deep Reinforcement Learning,” arXiv, arXiv:2006.12862, Feb. 2021.

\bibitem{c123} D. Yarats et al., “Reinforcement Learning with Prototypical Representations,” ICML, Jul. 2021, pp. 11920–11931.

\bibitem{c124} B. Amos et al., “On the Model-Based Stochastic Value Gradient for Continuous Reinforcement Learning,” L4DC, May 2021, pp. 6–20.

\bibitem{c18} D. Quillen et al., “Deep Reinforcement Learning for Vision-Based Robotic Grasping: A Simulated Comparative Evaluation of Off-Policy Methods,” ICRA, May 2018, pp. 6284–6291. doi: 10.1109/ICRA.2018.8461039.

\bibitem{c19} Y. Zhu et al., “robosuite: A Modular Simulation Framework and Benchmark for Robot Learning,” arXiv, arXiv:2009.12293, Sep. 2020. doi: 10.48550/arXiv.2009.12293.

\bibitem{c20} M. Ahn et al., “ROBEL: Robotics Benchmarks for Learning with Low-Cost Robots,” CoRL, May 2020, pp. 1300–1313. 

\bibitem{c21} A. Anand et al., “Unsupervised State Representation Learning in Atari,” NIPS, 2019, vol. 32.

\bibitem{c22} Y. Liang et al., “State of the Art Control of Atari Games Using Shallow Reinforcement Learning,” arXiv, arXiv:1512.01563, Apr. 2016. doi: 10.48550/arXiv.1512.01563.

\bibitem{c23} Abeyruwan, S. RLLib: Lightweight standard and on/off policy reinforcement learning library (C++). http://web.cs.miami.edu/home/saminda/rilib.html, 2013.

\bibitem{c24} Dutech et al., Reinforcement learning benchmarks and bake-offs ii. NIPS, 17, 2005.

\bibitem{c25} Dimitrakakis et al., The reinforcement learning competition 2014. AI Magazine, 35(3):61–65, 2014.

\bibitem{c26} J. Peters and S. Schaal, “Reinforcement learning by reward-weighted regression for operational space control,” ICML 2007, pp. 745–750. doi: 10.1145/1273496.1273590.

\bibitem{c27} C. C.-Y. Hsu et al., “Revisiting Design Choices in Proximal Policy Optimization,” arXiv, arXiv:2009.10897, Sep. 2020. doi: 10.48550/arXiv.2009.10897.


\bibitem{c28} S. Dasari et al., “RB2: Robotic Manipulation Benchmarking with a Twist,” Jan. 2022. Accessed: Aug. 18, 2022. [Online]. Available: https://openreview.net/forum?id=e82\_BlJL43M

\bibitem{c29} R. Kidambi et al., Morel: Model-based offline reinforcement learning. arXiv preprint arXiv:2005.05951, 2020.

\bibitem{c30} P. Aumjaud et al., “Reinforcement Learning Experiments and Benchmark for Solving Robotic Reaching Tasks,” Advances in Intelligent Systems and Computing, vol. 1285, pp. 318–331, Nov. 2020, doi: 10.1007/978-3-030-62579-5\_22.

\bibitem{c31} E. Todorov et al., “MuJoCo: A physics engine for model-based control,” in 2012 IEEE/RSJ International Conference on Intelligent Robots and Systems, Oct. 2012, pp. 5026–5033. doi: 10.1109/IROS.2012.6386109.

\bibitem{c32} Y. Tassa et al., “DeepMind Control Suite,” arXiv:1801.00690 [cs], Jan. 2018, Accessed: Apr. 25, 2022. [Online]. Available: http://arxiv.org/abs/1801.00690

\bibitem{c33} C. Devin et al., “Learning modular neural network policies for multi-task and multi-robot transfer,” in 2017 IEEE International Conference on Robotics and Automation (ICRA), May 2017, pp. 2169–2176. doi: 10.1109/ICRA.2017.7989250.

\bibitem{c34} A. Gupta et al., “Learning Invariant Feature Spaces to Transfer Skills with Reinforcement Learning,” arXiv, arXiv:1703.02949, Mar. 2017. doi: 10.48550/arXiv.1703.02949.

\bibitem{c35} M. Plappert et al., “Multi-Goal Reinforcement Learning: Challenging Robotics Environments and Request for Research,” arXiv, arXiv:1802.09464, Mar. 2018. doi: 10.48550/arXiv.1802.09464.

\bibitem{c36} T. Chen et al., “Hardware Conditioned Policies for Multi-Robot Transfer Learning,” arXiv, arXiv:1811.09864, Jan. 2019. doi: 10.48550/arXiv.1811.09864.

\bibitem{c37} A. Rupam Mahmood et al., “Setting up a Reinforcement Learning Task with a Real-World Robot,” in 2018 IEEE/RSJ International Conference on Intelligent Robots and Systems (IROS), Oct. 2018, pp. 4635–4640. doi: 10.1109/IROS.2018.8593894.

\bibitem{c38} S. Gu et al., “Deep Reinforcement Learning for Robotic Manipulation with Asynchronous Off-Policy Updates,” arXiv, arXiv:1610.00633, Nov. 2016. doi: 10.48550/arXiv.1610.00633.

\bibitem{c39} M. P. Deisenroth and C. E. Rasmussen, “PILCO: a model-based and data-efficient approach to policy search,” in Proceedings of the 28th International Conference on International Conference on Machine Learning, Madison, WI, USA, Jun. 2011, pp. 465–472.

\bibitem{c40} Y. Tassa et al., “dm\_control: Software and Tasks for Continuous Control,” arXiv:2006.12983 [cs], Sep. 2020, doi: 10.1016/j.simpa.2020.100022.

\bibitem{c41} T. Erez et al., “Simulation tools for model-based robotics: Comparison of Bullet, Havok, MuJoCo, ODE and PhysX,” in 2015 IEEE International Conference on Robotics and Automation (ICRA), May 2015, pp. 4397–4404. doi: 10.1109/ICRA.2015.7139807.

\bibitem{c42} R. BELLMAN, “A Markovian Decision Process,” Journal of Mathematics and Mechanics, vol. 6, no. 5, pp. 679–684, 1957.

\bibitem{c43} D. Yarats et al., “Improving Sample Efficiency in Model-Free Reinforcement Learning from Images,” Proceedings of the AAAI Conference on Artificial Intelligence, vol. 35, no. 12, Art. no. 12, May 2021.

\bibitem{c44} D. Hafner et al., “Mastering Atari with Discrete World Models,” arXiv, arXiv:2010.02193, Feb. 2022. doi: 10.48550/arXiv.2010.02193.

\bibitem{c45} P. Wawrzyński, “Control Policy with Autocorrelated Noise in Reinforcement Learning for Robotics,” International Journal of Machine Learning and Computing, vol. 5, no. 2, p. 91, 2015.

\bibitem{c46} W. Fedus et al., “Revisiting Fundamentals of Experience Replay,” in Proceedings of the 37th International Conference on Machine Learning, Nov. 2020, pp. 3061–3071. Accessed: Sep. 10, 2022.




\end{thebibliography}
\end{document}